\documentclass{article}

\usepackage{arxiv}

\usepackage[utf8]{inputenc} 
\usepackage[T1]{fontenc}    
\usepackage{hyperref}       
\usepackage{url}            
\usepackage{booktabs}       
\usepackage{amsfonts}       
\usepackage{nicefrac}       
\usepackage{microtype}      
\usepackage{lipsum}		
\usepackage{graphicx}
\usepackage{natbib}
\usepackage{doi}
\usepackage{amsmath}
\usepackage{tikz}
\usetikzlibrary{shapes, arrows.meta, positioning}
\usepackage{geometry}
\usepackage{tikz}
\usepackage{tikz}
\usetikzlibrary{positioning,arrows.meta,calc,fit}
\usepackage{xcolor}
\usetikzlibrary{mindmap, shadows, positioning, backgrounds}
\usetikzlibrary{positioning, shapes, arrows.meta}
\usepackage{amsmath, amssymb}

\title{Ultralytics YOLO Evolution: An Overview of YOLO27, YOLO26, YOLO11, YOLOv8, and YOLOv5 Object Detectors for Computer Vision and Pattern Recognition}

\author{
  \href{https://orcid.org/0000-0002-5417-6744}{\includegraphics[scale=0.06]{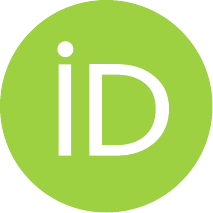}\hspace{1mm}Ranjan Sapkota\textsuperscript{}} \quad
  \href{https://orcid.org/0000-0001-5337-4848}{\includegraphics[scale=0.06]{orcid.pdf}\hspace{1mm}Manoj Karkee\textsuperscript{}} \\
  \textsuperscript{}Cornell University, Biological \& Environmental Engineering, Ithaca, NY 14850, USA \\
  \texttt{rs2672@cornell.edu} \\
}

\renewcommand{\shorttitle}{Ultralytics YOLO Evolution}

\hypersetup{
pdftitle={A template for the arxiv style},
pdfsubject={q-bio.NC, q-bio.QM},
pdfauthor={David S.~Hippocampus, Elias D.~Striatum},
pdfkeywords={First keyword, Second keyword, More},
}

\begin{document}
\maketitle

\begin{abstract}
This paper presents a comprehensive overview of the Ultralytics YOLO family, emphasizing architectural evolution, benchmarking, deployment, and emerging directions from YOLOv5 through YOLO27. The review begins with YOLO27 (or YOLOv27), which introduces a scale-adaptive dual-architecture strategy: compact YOLO27n/s detectors employ streamlined CNNs with dual-scale prediction, strengthened high-resolution features, foreground-alignment supervision, and conventional or NMS-free inference, whereas YOLO27m/l adopt query-based transformer decoding for native NMS-free detection. YOLO27l further incorporates an UltraViT backbone with deep-stage self-attention for global-context modeling. Preliminary COCO results span 42.3-60.4 mAP at 640-pixel resolution and 0.62-2.32 ms TensorRT 11 FP16 latency, with YOLO27l reaching 61.2 mAP at 800 pixels. The evolution is subsequently traced through YOLO26, including DFL removal, Progressive Loss Balancing, Small-Target-Aware Label Assignment, MuSGD optimization, and NMS-free inference; YOLO11, emphasizing efficiency and task integration; YOLOv8, introducing decoupled anchor-free detection; and YOLOv5, which established the modular PyTorch-based Ultralytics ecosystem. Comparative benchmarking examines accuracy, precision, recall, F1-score, mAP, latency, and computational complexity alongside representative contemporary detectors. The review further examines detection, segmentation, depth, classification, pose, oriented detection, tracking, export, quantization, and deployment across robotics, agriculture, surveillance, and manufacturing. Finally, challenges involving dense scenes, CNN-Transformer integration, open-vocabulary perception, domain generalization, and hardware-aware optimization are discussed as directions for future YOLO systems.
\end{abstract}


\section{Introduction}
\label{sec:introduction}

Object detection is one of the fundamental tasks in computer vision, enabling
machines to simultaneously identify and spatially localize multiple objects
within complex images and video streams
\cite{zhao2019object,zou2023object}. Its importance extends across autonomous
driving, robotics, surveillance, medical imaging, agriculture, and smart
manufacturing, where accurate and computationally efficient perception can
directly influence safety, productivity, autonomy, and real-time
decision-making \cite{rana2024artificial,khan2025objectdetection}. Among the
numerous detection frameworks developed over the past decade, the
\textit{You Only Look Once} (YOLO) paradigm has become one of the most
influential families of real-time object detectors by combining competitive
accuracy with high inference speed \cite{sapkota2025yolo}. Since its original
introduction in 2016, YOLO has undergone extensive architectural evolution,
incorporating advances in convolutional feature extraction, multi-scale
fusion, anchor design, task assignment, optimization, end-to-end prediction,
attention, and deployment-oriented inference \cite{sapkota2025yolo}. Within
the Ultralytics ecosystem, this progression is particularly evident from
YOLOv5 through YOLOv8, YOLO11, YOLO26, and the emerging YOLO27 generation.
As summarized in Figure~\ref{fig:yolotimeline}, this evolution reflects not
only increasing task breadth, but also a gradual transition from conventional
dense CNN-based prediction toward end-to-end and query-based detection
mechanisms.

\begin{figure*}[ht]
\centering

\begin{tikzpicture}[
    x=1cm,
    y=1cm,
    version/.style={
        rectangle,
        rounded corners=2pt,
        draw=black,
        very thick,
        minimum width=2.65cm,
        minimum height=1.05cm,
        align=center,
        font=\small\bfseries,
        fill=gray!8
    },
    latest/.style={
        version,
        draw=blue!65!black,
        line width=1.3pt,
        fill=blue!8
    },
    task/.style={
        rectangle,
        draw=black!65,
        rounded corners=2pt,
        minimum width=2.65cm,
        minimum height=0.62cm,
        align=center,
        font=\scriptsize,
        fill=white
    },
    newtask/.style={
        task,
        draw=blue!65!black,
        fill=blue!7,
        font=\scriptsize\bfseries
    },
    unsupported/.style={
        task,
        dashed,
        draw=gray!65,
        text=gray!70
    },
    architecture/.style={
        rectangle,
        rounded corners=2pt,
        draw=blue!65!black,
        minimum width=2.65cm,
        minimum height=0.70cm,
        align=center,
        font=\scriptsize\bfseries,
        fill=blue!5
    },
    arrowstyle/.style={
        -{Stealth[length=2.5mm]},
        thick,
        draw=black!75
    },
    evolve/.style={
        -{Stealth[length=3mm]},
        very thick,
        draw=blue!60!black
    }
]


\node[version] (yolo5) at (0,0)
{YOLOv5\\{\scriptsize 2020}};

\node[version] (yolo8) at (3.2,0)
{YOLOv8\\{\scriptsize 2023}};

\node[version] (yolo11) at (6.4,0)
{YOLO11\\{\scriptsize 2024}};

\node[version] (yolo26) at (9.6,0)
{YOLO26\\{\scriptsize 2026}};

\node[latest] (yolo27) at (12.8,0)
{YOLO27\\{\scriptsize 2026}};

\draw[evolve] (yolo5.east) -- (yolo8.west);
\draw[evolve] (yolo8.east) -- (yolo11.west);
\draw[evolve] (yolo11.east) -- (yolo26.west);
\draw[evolve] (yolo26.east) -- (yolo27.west);


\node[architecture, below=0.55cm of yolo5] (a5)
{Modular CNN\\Anchor-Based};

\node[architecture, below=0.55cm of yolo8] (a8)
{Anchor-Free CNN\\Decoupled Head};

\node[architecture, below=0.55cm of yolo11] (a11)
{Efficient CNN\\Multi-Task Design};

\node[architecture, below=0.55cm of yolo26] (a26)
{End-to-End Option\\NMS-Free Inference};

\node[architecture, below=0.55cm of yolo27] (a27)
{Scale-Adaptive\\Dual Architecture};


\node[task, below=0.65cm of a5] (y5det)
{Object Detection};

\node[task, below=0.18cm of y5det] (y5seg)
{Instance Segmentation*};

\node[task, below=0.18cm of y5seg] (y5cls)
{Classification*};

\node[unsupported, below=0.18cm of y5cls] (y5pose)
{Pose / Keypoints};

\node[unsupported, below=0.18cm of y5pose] (y5obb)
{Oriented Detection};

\node[unsupported, below=0.18cm of y5obb] (y5sem)
{Semantic Segmentation};

\node[unsupported, below=0.18cm of y5sem] (y5depth)
{Depth Estimation};


\node[task, below=0.65cm of a8] (y8det)
{Object Detection};

\node[task, below=0.18cm of y8det] (y8seg)
{Instance Segmentation};

\node[task, below=0.18cm of y8seg] (y8cls)
{Classification};

\node[task, below=0.18cm of y8cls] (y8pose)
{Pose / Keypoints};

\node[task, below=0.18cm of y8pose] (y8obb)
{Oriented Detection};

\node[unsupported, below=0.18cm of y8obb] (y8sem)
{Semantic Segmentation};

\node[unsupported, below=0.18cm of y8sem] (y8depth)
{Depth Estimation};


\node[task, below=0.65cm of a11] (y11det)
{Object Detection};

\node[task, below=0.18cm of y11det] (y11seg)
{Instance Segmentation};

\node[task, below=0.18cm of y11seg] (y11cls)
{Classification};

\node[task, below=0.18cm of y11cls] (y11pose)
{Pose / Keypoints};

\node[task, below=0.18cm of y11pose] (y11obb)
{Oriented Detection};

\node[unsupported, below=0.18cm of y11obb] (y11sem)
{Semantic Segmentation};

\node[unsupported, below=0.18cm of y11sem] (y11depth)
{Depth Estimation};


\node[task, below=0.65cm of a26] (y26det)
{Object Detection};

\node[task, below=0.18cm of y26det] (y26seg)
{Instance Segmentation};

\node[task, below=0.18cm of y26seg] (y26cls)
{Classification};

\node[task, below=0.18cm of y26cls] (y26pose)
{Pose / Keypoints};

\node[task, below=0.18cm of y26pose] (y26obb)
{Oriented Detection};

\node[unsupported, below=0.18cm of y26obb] (y26sem)
{Semantic Segmentation};

\node[unsupported, below=0.18cm of y26sem] (y26depth)
{Depth Estimation};


\node[newtask, below=0.65cm of a27] (y27det)
{Object Detection};

\node[newtask, below=0.18cm of y27det] (y27seg)
{Instance Segmentation};

\node[newtask, below=0.18cm of y27seg] (y27cls)
{Classification};

\node[newtask, below=0.18cm of y27cls] (y27pose)
{Pose / Keypoints};

\node[newtask, below=0.18cm of y27pose] (y27obb)
{Oriented Detection};

\node[newtask, below=0.18cm of y27obb] (y27sem)
{Semantic Segmentation};

\node[newtask, below=0.18cm of y27sem] (y27depth)
{Depth Estimation};


\node[
    rectangle,
    rounded corners=3pt,
    draw=blue!70!black,
    thick,
    fill=blue!5,
    minimum width=2.55cm,
    minimum height=0.85cm,
    align=center,
    font=\scriptsize\bfseries,
    below=0.70cm of y27depth,
    xshift=-1.45cm
] (y27ns)
{YOLO27n/s\\CNN Dual-Scale};

\node[
    rectangle,
    rounded corners=3pt,
    draw=orange!80!black,
    thick,
    fill=orange!7,
    minimum width=2.55cm,
    minimum height=0.85cm,
    align=center,
    font=\scriptsize\bfseries,
    below=0.70cm of y27depth,
    xshift=1.45cm
] (y27ml)
{YOLO27m/l\\Query-Based};

\coordinate (branch) at ($(y27depth.south)+(0,-0.32)$);

\draw[-, thick, draw=black!65]
(y27depth.south) -- (branch);

\draw[arrowstyle]
(branch) -| (y27ns.north);

\draw[arrowstyle]
(branch) -| (y27ml.north);


\draw[
    -{Stealth[length=3.5mm]},
    line width=1.2pt,
    draw=black!70
]
(-1.3,-8.2) -- (14.3,-8.2);

\node[
    font=\scriptsize\bfseries,
    fill=white,
    inner sep=3pt
] at (6.5,-8.2)
{Increasing Task Breadth, End-to-End Capability, and Architectural Specialization};


\node[
    task,
    minimum width=1.45cm,
    minimum height=0.45cm
] (legsolid) at (1.0,-9.0)
{Supported};

\node[
    unsupported,
    minimum width=1.45cm,
    minimum height=0.45cm
] (legdash) at (4.0,-9.0)
{Not Native};

\node[
    newtask,
    minimum width=1.65cm,
    minimum height=0.45cm
] (leg27) at (7.0,-9.0)
{YOLO27 Task};

\node[
    font=\scriptsize,
    anchor=west
] at (8.3,-9.0)
{*Latest models};

\end{tikzpicture}

\caption{
Evolution of task support and architectural design across representative
Ultralytics YOLO generations from YOLOv5 to YOLO27. YOLOv5 established the
modular PyTorch-based detection ecosystem, YOLOv8 introduced anchor-free
decoupled prediction, YOLO11 emphasized computational efficiency and
multi-task integration, and YOLO26 advanced end-to-end and NMS-free
inference. YOLO27 extends this progression through a scale-adaptive
dual-architecture strategy, employing streamlined CNN-based dual-scale
detection for YOLO27n/s and query-based transformer decoding for YOLO27m/l,
while expanding the planned vision-task ecosystem to segmentation, depth,
classification, pose, and oriented detection.
}
\label{fig:yolotimeline}
\end{figure*}

The conceptual foundation of this evolution originates from the original YOLO
formulation proposed by Redmon \textit{et al.}, which reframed object
detection as a unified regression problem that directly predicts bounding
boxes and class probabilities from an input image
\cite{redmon2016you}. This formulation represented a fundamental departure
from two-stage detectors such as R-CNN-derived architectures and Faster
R-CNN, where region proposal generation and subsequent object classification
are treated as distinct stages \cite{he2017mask,ren2016faster}. By eliminating
the explicit proposal stage, YOLOv1 demonstrated that object detection could
be formulated as a single forward prediction process, making the paradigm
particularly attractive for latency-sensitive applications such as robotics
and autonomous navigation \cite{ali2024yolo,diwan2023object}. Subsequent
generations progressively addressed limitations in localization accuracy,
small-object detection, scale variation, feature representation, and
computational efficiency.

YOLOv2 extended the original formulation through batch normalization, anchor
boxes, dimension clustering, and multi-scale training, improving robustness
across object sizes \cite{redmon2017yolo9000}. YOLOv3 subsequently introduced
the substantially deeper Darknet-53 backbone and multi-scale predictions,
establishing an influential balance between computational efficiency and
detection accuracy
\cite{redmon2018yolov3,apostolidis2025delving,sapkota2025yolo,
edozie2025comprehensive}. As application requirements expanded toward aerial
imagery, agriculture, medical analysis, industrial inspection, and other
challenging environments, the YOLO architecture diversified further. YOLOv4
combined CSPNet-based feature extraction with improved activation functions,
feature aggregation, mosaic augmentation, Complete Intersection over Union
(CIoU), and other ``bag-of-freebies'' and ``bag-of-specials'' strategies
\cite{bochkovskiy2020yolov4}. These developments established several
principles that remained influential in later YOLO systems: hierarchical CNN
feature extraction, multi-resolution feature fusion, dense prediction, and
joint optimization of accuracy and latency.

Within this broader evolution, Ultralytics YOLOv5 established a modular,
PyTorch-oriented implementation and deployment ecosystem that substantially
simplified model training, customization, validation, and export. Its
engineering-oriented design helped expand YOLO from a detector architecture
into a reusable computer-vision framework supporting practical deployment
across heterogeneous hardware. Parallel developments such as YOLOv6 and
YOLOv7 introduced efficient architectural components, model
reparameterization, and improved training strategies to further strengthen
real-time performance \cite{li2022yolov6,wang2023yolov7}. Ultralytics later
introduced YOLOv8 with a decoupled detection head and anchor-free prediction
strategy, reducing dependence on predefined anchor configurations while
providing a more unified foundation for multiple computer-vision tasks
\cite{sohan2024review,farooq2024improved}. The evolution of these capabilities
and their expansion across detection, segmentation, classification, pose,
and oriented detection are illustrated in Figure~\ref{fig:yolotimeline}.

Further developments increasingly challenged assumptions inherited from
classical dense YOLO detectors. YOLOv9 explored Generalized Efficient Layer
Aggregation Network (GELAN)-based representations and Programmable Gradient
Information to improve information preservation during optimization
\cite{wang2024yolov9}, whereas YOLOv10 advanced end-to-end detection through
training assignments designed to reduce dependence on conventional
post-processing \cite{wang2024yolov10}. YOLO11 continued the Ultralytics
lineage with an emphasis on computational efficiency, task integration, and
strong real-time performance \cite{sapkota2025yolo}. In parallel,
non-Ultralytics YOLO-named architectures explored alternative design
directions. YOLOv12 investigated attention-centric real-time detection, while
YOLOv13 introduced higher-order feature interactions and enhanced relational
representation \cite{tian2025yolov12,lei2025yolov13}. Collectively, these
developments indicate that modern YOLO research has increasingly moved beyond
simply scaling convolutional backbones toward redesigning feature
interaction, assignment, inference, and global-context reasoning.

A particularly important milestone in the Ultralytics trajectory is YOLO26,
which advances a deployment-oriented formulation through architectural and
training simplifications. YOLO26 removes Distribution Focal Loss (DFL),
supports native end-to-end NMS-free inference, and incorporates Progressive
Loss Balancing (ProgLoss), Small-Target-Aware Label Assignment (STAL), and the
MuSGD optimizer to improve optimization stability and deployment efficiency
\cite{sapkota2025yolo26}. These modifications address several practical
limitations associated with dense detection pipelines, including
post-processing overhead, optimization imbalance, small-target assignment,
and export complexity. Consequently, YOLO26 represents an important
transitional stage between conventional dense CNN-based YOLO detectors and
more explicitly end-to-end detection paradigms. Rather than viewing NMS as an
unavoidable final stage, this generation strengthens the possibility of
direct prediction pipelines in which training objectives are more closely
aligned with inference behavior.

YOLO27 (or YOLOv27) extends this progression through a more fundamental
architectural departure: instead of assuming that a single detector
architecture should optimally serve every model scale, YOLO27 introduces a
\textit{scale-adaptive dual-architecture} strategy. The compact YOLO27n and
YOLO27s detection models retain an efficiency-oriented CNN pathway, whereas
the higher-capacity YOLO27m and YOLO27l detectors transition toward
query-based prediction with transformer decoding. Thus, model scaling in
YOLO27 is not restricted to increasing network depth, width, or parameter
count; it can also alter the internal detection paradigm itself. This
distinction is highlighted in Figure~\ref{fig:yolotimeline}, where the
YOLO27 family separates into compact CNN-based n/s models and higher-capacity
query-based m/l models, representing a transition from predominantly uniform
architectural scaling toward scale-dependent specialization.

For YOLO27n/s, the compact detection pathway is designed around a streamlined
CNN architecture with dual-scale prediction. Rather than retaining the
conventional three-level detection hierarchy, these models remove the
intermediate detection scale while strengthening high-resolution feature
extraction. This design seeks to allocate representational capacity more
efficiently toward fine spatial features relevant to small-object perception
while avoiding unnecessary computational overhead at the intermediate scale.
The compact models further introduce foreground-alignment supervision during
training to reduce the discrepancy between dense one-to-many supervision and
one-to-one prediction. Importantly, they can retain conventional dense
prediction followed by Non-Maximum Suppression (NMS), while also supporting
a one-to-one NMS-free inference mode. This dual prediction behavior provides
a practical bridge between established dense YOLO inference and emerging
end-to-end formulations.

YOLO27m and YOLO27l represent the more substantial architectural transition.
Rather than relying exclusively on dense convolutional detection heads, these
models employ multi-scale visual features together with object queries and
transformer decoding to directly generate class and bounding-box predictions.
The query-based formulation enables native NMS-free inference by treating
detection as a set-prediction problem rather than producing large collections
of overlapping candidate boxes that must subsequently be filtered. YOLO27m
combines CNN-based feature extraction with the query decoder, providing a
hybrid pathway between conventional convolutional perception and
transformer-based detection. YOLO27l extends this concept through the
UltraViT backbone, in which convolution remains important in shallow and
intermediate stages while self-attention is introduced at deeper,
lower-resolution semantic stages. This arrangement is intended to strengthen
global-context modeling without imposing the full computational cost of
attention across high-resolution feature maps. Consequently, YOLO27m/l
represent a convergence between the efficiency principles historically
associated with YOLO and the set-based reasoning mechanisms established by
query-driven transformer detectors.

The significance of YOLO27 therefore extends beyond a conventional
incremental model update. Its architecture suggests that the future evolution
of real-time detectors may increasingly depend on \textit{architectural
specialization by deployment scale}. Compact models operating under stringent
memory, latency, and energy constraints may continue to benefit from
optimized convolutional representations, whereas larger models deployed on
more capable accelerators can exploit query-based decoding and global-context
reasoning. This strategy also challenges the conventional assumption that
nano, small, medium, and large variants should simply represent uniformly
scaled versions of the same computational graph. Instead, YOLO27 treats
computational budget as an architectural design variable.

Preliminary results provide an initial quantitative indication of this
trade-off. At 640-pixel resolution, the YOLO27 detection family reports COCO
mAP$_{50:95}$ values ranging from 42.3 for YOLO27n to 60.4 for YOLO27l,
with TensorRT~11 FP16 GPU inference latency ranging from approximately
0.62~ms to 2.32~ms on an NVIDIA RTX PRO 6000. YOLO27l additionally reaches
61.2 mAP at 800-pixel resolution. These preliminary measurements suggest that
the transition from compact CNN-based models to higher-capacity query-based
models can substantially increase detection accuracy while retaining
real-time inference characteristics. Nevertheless, such results should be
interpreted as preliminary until final model weights, configurations,
benchmarking protocols, and independent evaluations are available.

The YOLO27 ecosystem also extends beyond conventional bounding-box detection.
The planned model family encompasses instance segmentation, semantic
segmentation, monocular depth estimation, image classification, human pose
estimation, oriented bounding-box detection, and tracking-compatible
inference. Importantly, this broader task ecosystem does not necessarily
adopt the same query-based formulation used by the medium and large detection
models; non-detection tasks retain CNN-oriented task-specific architectures.
This distinction reinforces the central design philosophy of YOLO27:
architectural choices are increasingly determined by model scale, task
requirements, and deployment constraints rather than by enforcing a single
internal architecture across the entire family.

The objective of this paper is to provide a consolidated perspective on
how Ultralytics YOLO has progressed from modular CNN-based dense detection
toward anchor-free prediction, end-to-end inference, and ultimately
scale-adaptive CNN--query architectures. Particular attention is given to
architectural innovations, accuracy--efficiency trade-offs, benchmarking,
multi-task capability, export and deployment considerations, and practical
model selection. Cross-comparisons with related YOLO generations and
contemporary real-time detectors further contextualize these developments.
Through this perspective, the paper characterizes YOLO evolution not merely
as a sequence of increasingly accurate models, but as a broader transition
toward task-aware, hardware-aware, and scale-dependent real-time perception
systems for computer vision, robotics, autonomous systems, agriculture,
surveillance, manufacturing, and edge AI.

\section{Architectural Evolution of Ultralytics YOLO Models}

To provide a precise chronological and architectural context, we separate Ultralytics-maintained releases from community-driven variants. \textit{Table~\ref{tab:ultralytics}} catalogs the Ultralytics lineage YOLOv5 (2020), YOLOv8 (2023), YOLO11 (2024), and YOLO26 (2025) highlighting design choices, capabilities, and deployment characteristics within a single vendor family. In contrast, \textit{Table~\ref{tab:nonultralytics}} surveys major community-driven releases independent of Ultralytics, including YOLOv1 (2015), YOLOv2 (2016), YOLOv3 (2018), YOLOv4 (2020), YOLOv6 (2022), YOLOv7 (2022), YOLOv9 (2024), and YOLOv10 (2024), alongside other contemporary variants. Together, these tables present a coherent, year-by-year view of architectural innovations, task expansions, and performance trends across the YOLO ecosystem, clarifying how the framework matured into the state represented by YOLO26 (Tables~\ref{tab:ultralytics}-\ref{tab:nonultralytics}).

\begin{table}[h!]
\centering
\caption{Ultralytics YOLO models: key architectural innovations, supported tasks, and implementation frameworks.}
\label{tab:ultralytics}
\scriptsize
\renewcommand{\arraystretch}{1.15}
\setlength{\tabcolsep}{3pt}

\begin{tabular}{p{2.1cm} p{9.3cm} p{3.4cm} p{1.3cm}}
\hline
\textbf{Model (Year)} &
\textbf{Key Architectural Innovation and Contribution} &
\textbf{Tasks} &
\textbf{Framework} \\
\hline

YOLOv5 (2020)
(\href{https://github.com/ultralytics/yolov5}{Source Link})
&
Established Ultralytics' PyTorch-based YOLO implementation and modular
training ecosystem; employed CSP-inspired convolutional feature extraction,
PANet-style multi-scale feature aggregation, SiLU activations, and
anchor-based detection. Its accessible training, augmentation, validation,
export, and deployment pipeline substantially broadened practical adoption
of YOLO across research and industrial applications.
&
Object Detection, Instance Segmentation, Classification
&
PyTorch
\\

YOLOv8 (2023)
(\href{https://docs.ultralytics.com/models/yolov8/}{Source Link})
&
Introduced a redesigned Ultralytics architecture with C2f modules,
anchor-free prediction, and a decoupled detection head, improving feature
representation and optimization while simplifying bounding-box prediction.
It also strengthened the unified Ultralytics interface for multiple
computer-vision tasks and deployment targets.
&
Object Detection, Instance Segmentation, Classification, Pose Estimation,
Oriented Detection
&
PyTorch
\\

YOLO11 (2024)
(\href{https://docs.ultralytics.com/models/yolo11/}{Source Link})
&
Advanced computational efficiency through C3k2 CSP-style bottlenecks and
C2PSA spatial-attention modules, improving feature extraction while
maintaining real-time inference. It further consolidated detection,
segmentation, classification, pose estimation, and oriented bounding-box
prediction within the unified Ultralytics ecosystem.
&
Object Detection, Instance Segmentation, Classification, Pose Estimation,
Oriented Detection
&
PyTorch
\\

YOLO26 (2026)
(\href{https://docs.ultralytics.com/models/yolo26/}{Source Link})
&
Advanced deployment-oriented YOLO design through native end-to-end,
NMS-free inference and removal of Distribution Focal Loss (DFL), simplifying
bounding-box regression and model export. It introduced Progressive Loss
Balancing (ProgLoss), Small-Target-Aware Label Assignment (STAL), and the
MuSGD optimizer to improve training stability, small-object learning, and
deployment efficiency.
&
Object Detection, Instance Segmentation, Classification, Pose Estimation,
Oriented Detection
&
PyTorch
\\

YOLO27 (2026)
(\href{https://docs.ultralytics.com/models/yolo27}{Source Link})
&
Introduces a \textbf{scale-adaptive dual-architecture} strategy rather than
using one detector architecture across all model scales. YOLO27n/s employ
streamlined CNNs with strengthened high-resolution features, dual-scale
prediction, and foreground-alignment supervision, supporting dense
one-to-many detection with NMS or one-to-one NMS-free inference.
YOLO27m/l instead adopt query-based transformer decoding for native NMS-free
detection, while YOLO27l further incorporates the UltraViT backbone with
deep-stage self-attention for enhanced global-context modeling.
&
Object Detection, Instance Segmentation, Semantic Segmentation, Depth
Estimation, Classification, Pose Estimation, Oriented Detection, Tracking
&
PyTorch
\\

\hline
\end{tabular}
\end{table}

\textbf{Non-Ultralytics YOLO versions:}

\begin{table}[h!]\centering
\caption{Other YOLO models(non-ultralytics): key architectural innovations, tasks, and frameworks}
\label{tab:nonultralytics}
\scriptsize
\begin{tabular}{p{2.5cm} p{8.5cm} p{3.2cm} p{1.5cm}}
\hline
\textbf{Model (Year)} & \textbf{Key Architectural Innovation and Contribution} & \textbf{Tasks} & \textbf{Framework} \\
\hline
YOLOv1 (2015) \cite{redmon2016you}& First unified single-stage detector; predicted bounding boxes + class probabilities in a single forward pass, pioneering real-time object detection. & Object Detection, Classification & Darknet \\
YOLOv2 (2016) \cite{redmon2017yolo9000} & Introduced anchor boxes and dimension clustering; enabled multi-scale training; YOLO9000 demonstrated joint detection and classification across 9000 categories. & Object Detection, Classification & Darknet \\
YOLOv3 (2018) \cite{redmon2018yolov3} & Deeper Darknet-53 backbone with residual blocks; multi-scale predictions enhanced small-object detection; added SPP for richer feature context. & Object Detection, Multi-scale Detection & Darknet \\
YOLOv4 (2020)  \cite{bochkovskiy2020yolov4} & CSPDarknet-53 backbone and Mish activation; improved training stability; optimized for high accuracy-speed tradeoff on GPUs. & Object Detection, Object Tracking & Darknet \\
YOLOv6 (2022)  \cite{li2022yolov6}& Industrial-focused release (by Meituan): EfficientRep backbone, improved self-attention, anchor-free head; focused on deployment efficiency for production. & Object Detection, Instance Segmentation & PyTorch \\
YOLOv7 (2022) \cite{wang2023yolov7} & Introduced E-ELAN for extended feature learning; re-parameterized convolution modules; incorporated transformer blocks for context modeling; optimized for tracking and multi-task vision. & Object Detection, Tracking, Segmentation & PyTorch \\
YOLOv9 (2024) \cite{wang2024yolov9} & Introduced Programmable Gradient Information (PGI) for enhanced learning and G-ELAN backbone; boosted training flexibility and feature richness. & Object Detection, Instance Segmentation & PyTorch \\
YOLOv10 (2024) \cite{wang2024yolov10} & Achieved end-to-end NMS-free detection using consistent dual-assignment training strategy; removed reliance on post-processing. & Object Detection & PyTorch \\
YOLOv12 (2025) \cite{tian2025yolov12} & Attention-centric design: efficient area attention module and Residual ELAN (R-ELAN) blocks; achieved transformer-level accuracy at YOLO speed. & Object Detection & PyTorch \\
YOLOv13 (2025) \cite{lei2025yolov13} & Proposed HyperACE (hypergraph-based correlation enhancement) and FullPAD (pipeline-wide aggregation/distribution) for global high-order feature learning; depthwise separable convolutions improved efficiency. & Object Detection & PyTorch \\
\hline
\end{tabular}
\end{table}

Ultralytics-maintained YOLO releases illustrate a steady trajectory toward modularity, task unification, and deployment efficiency, culminating in YOLO26 as the first fully integrated framework for detection, segmentation, pose estimation, oriented bounding boxes, and classification. In contrast, community-driven models have often prioritized experimental architectural features ranging from Darknet-based foundations in YOLOv1–YOLOv4 to transformer-inspired mechanisms and attention modules in YOLOv6–YOLOv13. While these variants advanced accuracy benchmarks and introduced novel ideas such as GELAN or hypergraph-based feature aggregation, they frequently retained computationally demanding components such as NMS or DFL, which limited seamless edge deployment. Ultralytics models, by contrast, consistently emphasize exportability, quantization, and production-ready APIs, reflecting a pragmatic design philosophy that complements rather than competes with academic prototypes. The comparative overview in Table~\ref{tab:nonultralytics} underscores how Ultralytics’ streamlined design choices have positioned YOLO26 as both a research-relevant and industry-ready milestone within the broader YOLO ecosystem. 

\subsection{YOLO27: Scale-Adaptive Dual Architecture, Dual-Scale CNNs, and Query-Based Detection}
\label{subsec:yolo27}

\begin{figure}[h!]
     \centering
     \includegraphics[width=\linewidth]{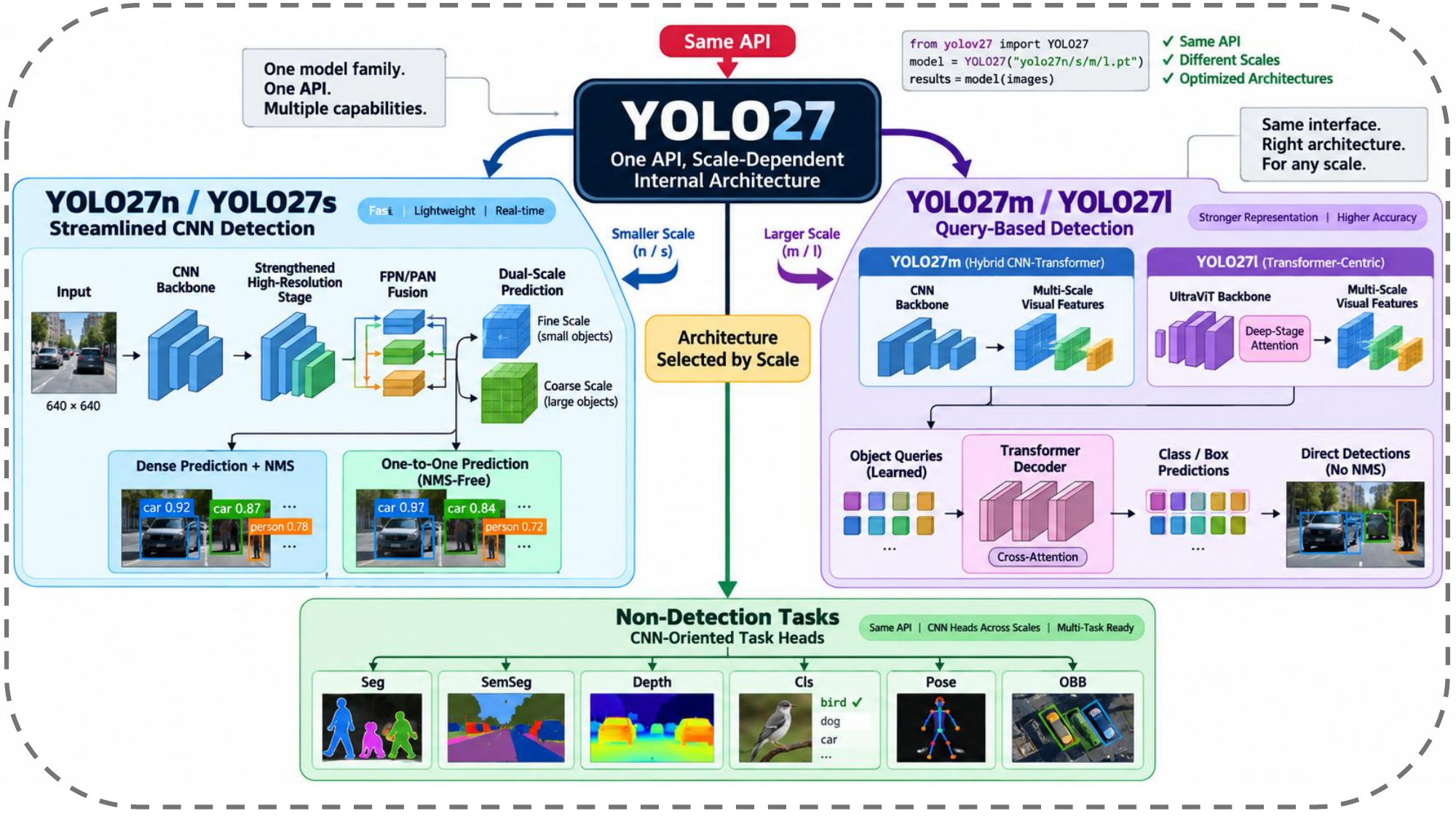}
    \caption{YOLO27 scale-adaptive dual-architecture. A unified API dynamically
    supports scale-dependent internal designs: YOLO27n/s employ streamlined
    CNN backbones, strengthened high-resolution features, dual-scale prediction,
    and either NMS-based or one-to-one NMS-free inference, whereas YOLO27m/l
    use query-based transformer decoding, with UltraViT attention in YOLO27l.
    Non-detection tasks retain CNN-oriented heads for segmentation, depth,
    classification, pose, and OBB prediction.}
    \label{fig:FIG2}
\end{figure}

YOLO27 (or YOLOv27) represents a substantial architectural transition in the
Ultralytics YOLO ecosystem by challenging the conventional assumption that
nano, small, medium, and large variants should be scaled realizations of a
single detector architecture
(\href{https://docs.ultralytics.com/models/yolo27}{Ultralytics YOLO27}).
Instead, YOLO27 introduces a \textbf{scale-adaptive dual-architecture}
strategy in which the internal detection paradigm changes according to model
capacity. As illustrated in Figure~\ref{fig:FIG2}, compact YOLO27n and
YOLO27s models retain an efficiency-oriented convolutional neural network
(CNN) pathway, whereas the higher-capacity YOLO27m and YOLO27l models adopt
query-based detection with transformer decoding. This design establishes a
clear architectural distinction between models intended for constrained
real-time deployment and models capable of exploiting greater computational
capacity for richer contextual reasoning.

For the compact \textbf{YOLO27n/s} variants, the architecture employs a
streamlined CNN backbone coupled with strengthened high-resolution feature
extraction and \textbf{dual-scale detection}. In contrast to conventional
three-level YOLO detection hierarchies, YOLO27n/s remove the intermediate
prediction scale and concentrate prediction on fine- and coarse-resolution
feature levels. The strengthened high-resolution pathway is intended to
preserve spatial detail required for small and localized objects, while the
coarser feature level retains stronger semantic information for larger
instances. This redistribution of computational capacity is particularly
relevant to latency-sensitive applications in which maintaining multiple
prediction scales may introduce unnecessary memory movement and computational
overhead. Consequently, dual-scale prediction should be interpreted not
simply as removal of a detection level, but as a reallocation of feature
capacity toward the resolutions considered most useful for compact real-time
detectors.

A second important component of the compact architecture is
\textbf{foreground-alignment supervision}. Dense YOLO detectors traditionally
benefit from one-to-many training assignments, where multiple candidate
locations can contribute supervisory signals for a single ground-truth
instance. End-to-end inference, however, favors one-to-one predictions that
produce a compact and non-redundant set of detections. This difference can
create a discrepancy between the optimization behavior that benefits dense
training and the prediction behavior required for NMS-free inference.
YOLO27 addresses this issue by introducing training-time foreground alignment
to improve consistency between these regimes. As summarized in
Figure~\ref{fig:FIG2}, YOLO27n/s therefore retain two practical inference
possibilities: a conventional dense one-to-many prediction pathway followed
by \textbf{Non-Maximum Suppression (NMS)}, and an optional one-to-one pathway
for \textbf{NMS-free inference}. This flexibility allows compact models to
preserve the established dense YOLO paradigm while simultaneously supporting
a more direct end-to-end deployment configuration.

The higher-capacity \textbf{YOLO27m/l} variants depart more fundamentally from
traditional dense YOLO prediction. Rather than generating a large collection
of candidate boxes that must subsequently be suppressed, these models employ
\textbf{object queries and transformer-based decoding} to directly predict a
compact set of object classes and bounding boxes. Multi-scale visual features
generated by the backbone and feature-fusion network provide the spatial and
semantic representation consumed by learned object queries. The transformer
decoder then iteratively associates these queries with relevant image
features, producing final set-based detections. Because the resulting
prediction formulation is explicitly designed to avoid redundant detections,
YOLO27m/l provide \textbf{native NMS-free inference}. This transition brings
the YOLO family closer to query-based end-to-end detection while retaining
the real-time efficiency principles that have historically characterized YOLO
architectures.

Importantly, the medium and large variants do not employ identical feature
extractors. \textbf{YOLO27m} retains a CNN-oriented backbone derived from the
preceding Ultralytics design and couples it with the query-based transformer
decoder, effectively forming a hybrid CNN--query architecture. In contrast,
\textbf{YOLO27l} introduces the \textbf{UltraViT} backbone, which combines
convolutional feature extraction with self-attention at deeper semantic
stages. Convolution remains advantageous at shallow and intermediate
resolutions because of its strong local inductive bias and computational
efficiency, whereas self-attention is introduced at deeper, lower-resolution
stages where global relationships can be modeled at substantially lower
computational cost. This arrangement enables YOLO27l to strengthen
\textbf{global-context modeling} without applying expensive attention
operations uniformly across high-resolution feature maps. The architectural
division between these pathways is explicitly visualized in
Figure~\ref{fig:FIG2}.

The architectural specialization of YOLO27 is accompanied by promising
preliminary accuracy--efficiency characteristics. At an input resolution of
640 pixels, the reported COCO mAP$_{50:95}$ increases from \textbf{42.3} for
YOLO27n to \textbf{49.6}, \textbf{55.8}, and \textbf{60.4} for YOLO27s,
YOLO27m, and YOLO27l, respectively. Corresponding TensorRT~11 FP16 inference
latencies on an NVIDIA RTX PRO 6000 are approximately \textbf{0.62},
\textbf{0.79}, \textbf{1.39}, and \textbf{2.32 ms}. Model capacity similarly
scales from approximately 3.0 million parameters for YOLO27n to 72.3 million
for YOLO27l. At an increased input resolution of 800 pixels, YOLO27l is
reported to reach \textbf{61.2 mAP}. These preliminary results indicate a
clear accuracy--latency continuum across the family, with compact CNN variants
targeting stringent latency budgets and query-based variants allocating
additional computational capacity toward detection accuracy and contextual
reasoning
(\href{https://docs.ultralytics.com/models/yolo27}{Ultralytics YOLO27}).
Because YOLO27 remains an emerging model family, however, these values should
be treated as preliminary until final weights, configurations, and independent
benchmarking become available.

Beyond bounding-box detection, YOLO27 broadens the Ultralytics perception
ecosystem toward \textbf{instance segmentation, semantic segmentation, depth
estimation, image classification, pose estimation, oriented bounding-box
(OBB) detection, and tracking-compatible inference}. Importantly, the
query-based architecture of YOLO27m/l detection should not be generalized to
all of these tasks. As shown in Figure~\ref{fig:FIG2}, non-detection tasks
retain CNN-oriented architectures and task-specific prediction heads. This
separation indicates that YOLO27 is not simply replacing convolutional
prediction with transformers; instead, it selectively introduces query-based
reasoning where its set-prediction properties are advantageous while retaining
efficient convolutional computation for other vision objectives.

From a deployment perspective, this design creates a more explicit
\textbf{accuracy--compute specialization} across model scales. YOLO27n/s are
naturally positioned for embedded perception, UAVs, mobile robotics, smart
cameras, and other resource-constrained systems where memory consumption and
batch-one latency are critical. YOLO27m/l provide a higher-capacity pathway
for GPU-equipped robotic systems, autonomous platforms, industrial vision,
and accuracy-critical perception workloads. The optional NMS-free pathway in
compact models and native NMS-free query decoding in larger models can also
simplify deployment pipelines by reducing dependence on scenario-specific
post-processing. Nevertheless, actual deployment advantages remain
hardware-dependent and require direct evaluation of latency, memory,
throughput, energy consumption, and quantization behavior on the intended
target platform.

Overall, YOLO27 represents a conceptual shift from \emph{uniform architectural
scaling} toward \emph{scale-dependent architectural specialization}. As
illustrated in Figure~\ref{fig:FIG2}, the family combines efficient
dual-scale CNN detection for YOLO27n/s with query-based transformer decoding
for YOLO27m/l, while UltraViT further introduces deep-stage self-attention in
YOLO27l. This organization suggests a broader direction for future real-time
vision systems: compact models may continue to exploit the computational
advantages of convolution, whereas larger variants increasingly integrate
queries, attention, and global reasoning. YOLO27 therefore extends the YOLO
paradigm beyond conventional backbone scaling toward a heterogeneous,
task-aware, and deployment-aware architecture designed to span edge AI,
real-time robotics, and high-accuracy GPU perception.

\subsection{YOLO26: DFL Removal, NMS-Free Inference, ProgLoss, STAL, MuSGD}
YOLO26 (2025) is an edge-first redesign that codifies end-to-end simplicity and export robustness. Two decisive architectural changes lead this release. First, the head removes Distribution Focal Loss (DFL)–based distributional regression in favor of a lighter, hardware-friendly parameterization of bounding boxes. Eliminating DFL prunes operators that were brittle across compilers and runtimes, reducing graph complexity and easing quantization. Second, the decoding path is reworked for \textbf{NMS-free, end-to-end inference}: the head produces a compact, non-redundant set of predictions without a post-processing suppression step. This erases a traditional latency bottleneck and removes deployment-time hyperparameters (IoU/score thresholds) that often demand scenario-specific tuning.

\begin{figure}[h!]
     \centering
     \includegraphics[width=0.89\linewidth]{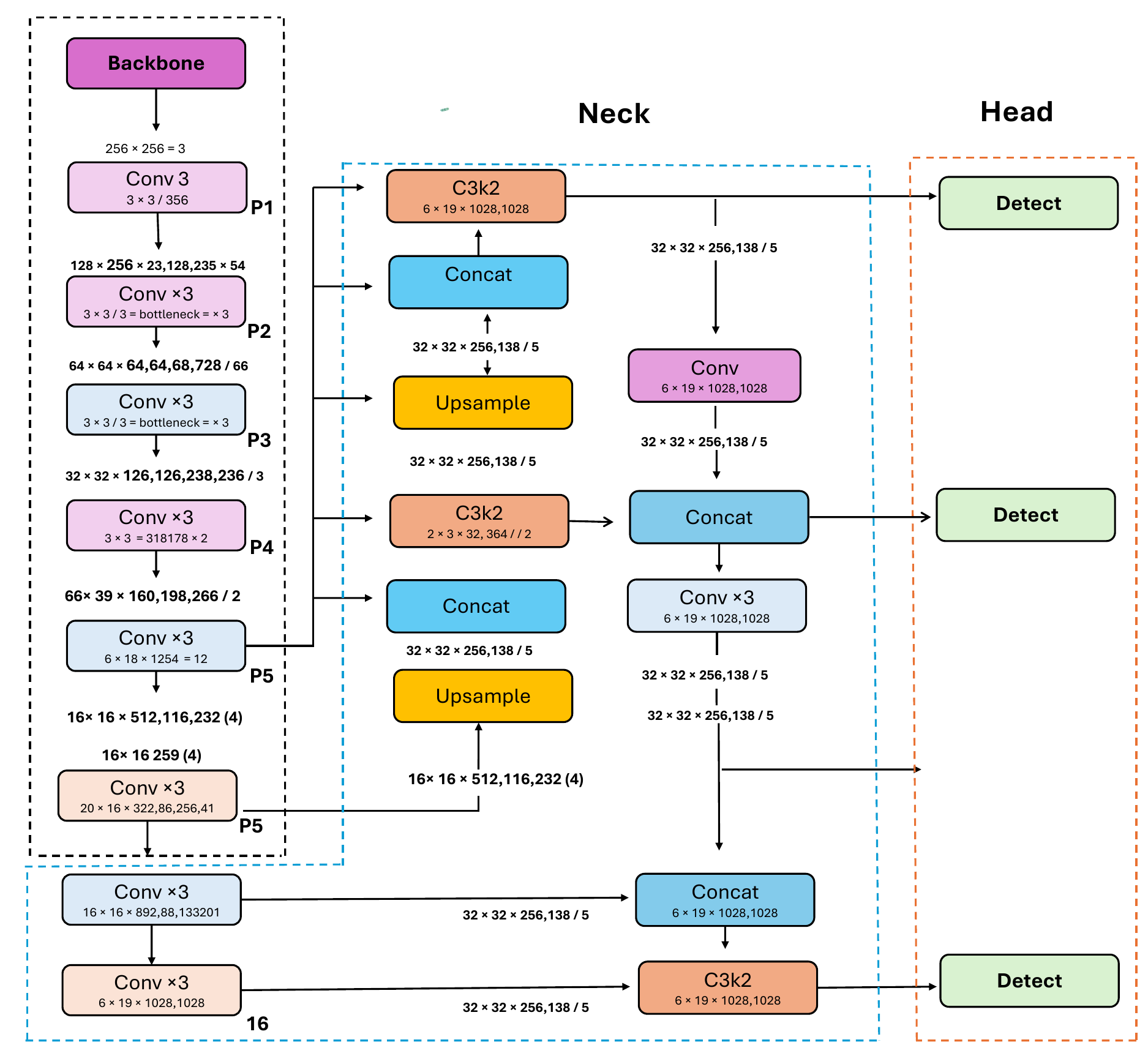}
    \caption{YOLO26 core architecture diagram}
    \label{fig:yolo26}
\end{figure}

Training dynamics are stabilized by \textbf{ProgLoss} (progressive loss balancing) and \textbf{STAL} (small-target–aware label assignment). ProgLoss schedules the relative weights of classification, localization, and auxiliary terms to prevent domination by easy negatives or large objects in later epochs, improving convergence smoothness. STAL adjusts assignment priors and spatial tolerance to ensure tiny, occluded, or low-contrast instances receive adequate supervisory signal, boosting recall in edge-centric imagery (UAV, smart cameras, mobile robotics). Optimization is driven by \textbf{MuSGD}, a hybrid that combines the generalization and simplicity of SGD with curvature- and momentum-aware updates inspired by modern large-model training. Empirically, MuSGD shortens time-to-quality and mitigates late-epoch oscillations, aiding reproducible convergence across scales (n/s/m/l/x).

On the systems side, YOLO26 leans into \textbf{portability and quantization}. By removing DFL and NMS, the exported graphs map cleanly to ONNX, TensorRT, CoreML, and TFLite with fewer custom kernels, facilitating INT8 and FP16 deployment with minimal accuracy loss. The simplified operator surface reduces conversion friction and improves determinism across compilers. Latency benefits accrue on CPUs and embedded GPUs, with notable gains at batch size one where post-processing previously dominated. With unified support for detection, instance segmentation, classification, pose/keypoints, and oriented bounding boxes, YOLO26 consolidates the family into a single, multi-task framework. In effect, it operationalizes a long-standing objective for real-time detection: align state-of-the-art accuracy with a \emph{minimal} and \emph{export-friendly} graph that scales from cloud GPUs to resource-constrained edge devices.

YOLO26 architecture (Figure \ref{fig:yolo26}) consolidates deployability and accuracy with two core simplifications: removing Distribution Focal Loss to ease export/hardware compatibility and enabling native end-to-end (NMS-free) inference to eliminate post-processing latency. Training stability and small-object recall are strengthened via Progressive Loss Balancing and Small-Target-Aware Label assignment, while MuSGD blends SGD with Muon-style curvature/momentum for fast, smooth convergence. Early previews indicate substantial CPU-side speedups ( approximately 43\%) and consistent performance under FP16/INT8 quantization, making the family well suited for edge devices (Jetson/CPU) alongside standard GPU pipelines.

\subsection{YOLO11: Hybrid Task Assignment and Efficiency}
YOLOv11 (2024) focused on balancing accuracy, stability, and device efficiency with refinements that touched both representation and supervision. On the representation side, compact CSP variants (e.g., C3/C3k2 bottlenecks) and lightweight attention modules were emphasized to increase channel interaction without inflating latency. These changes improved feature reuse and reduced redundancy, yielding better FLOPs-to-mAP ratios on mainstream GPUs and edge SoCs. Neck design continued to target low-loss multi-scale fusion with careful stride harmonization, limiting aliasing and improving localization at small scales.

The supervisory signal adopted a hybrid, task-aware assignment: label assignment and loss weighting were tuned jointly across classification, localization, and (where enabled) auxiliary tasks such as segmentation or pose. This hybridization reduced failure modes in crowded scenes and improved recall at fixed precision by stabilizing which features are optimized for which objective at each training stage. Importantly, the assignment strategy was designed to be robust to batch-size variability, an increasingly common constraint in edge training and fine-tuning scenarios.

From a system perspective, YOLOv11 strengthened multi-task support in the Ultralytics stack object detection, instance segmentation, classification, pose estimation, and (in supported builds) oriented bounding boxes while preserving the same export surfaces introduced in prior releases. It retained NMS-based decoding but reduced post-processing load through crisper, less redundant predictions. Together, the efficiency gains and hybrid task assignment positioned YOLOv11 as a versatile mid-generation release: more accurate than YOLOv8 under similar latency budgets, and more stable under transfer and domain shift, particularly for small-object and moderately dense scenes.

\begin{figure}[h!]
     \centering
     \includegraphics[width=0.99\linewidth]{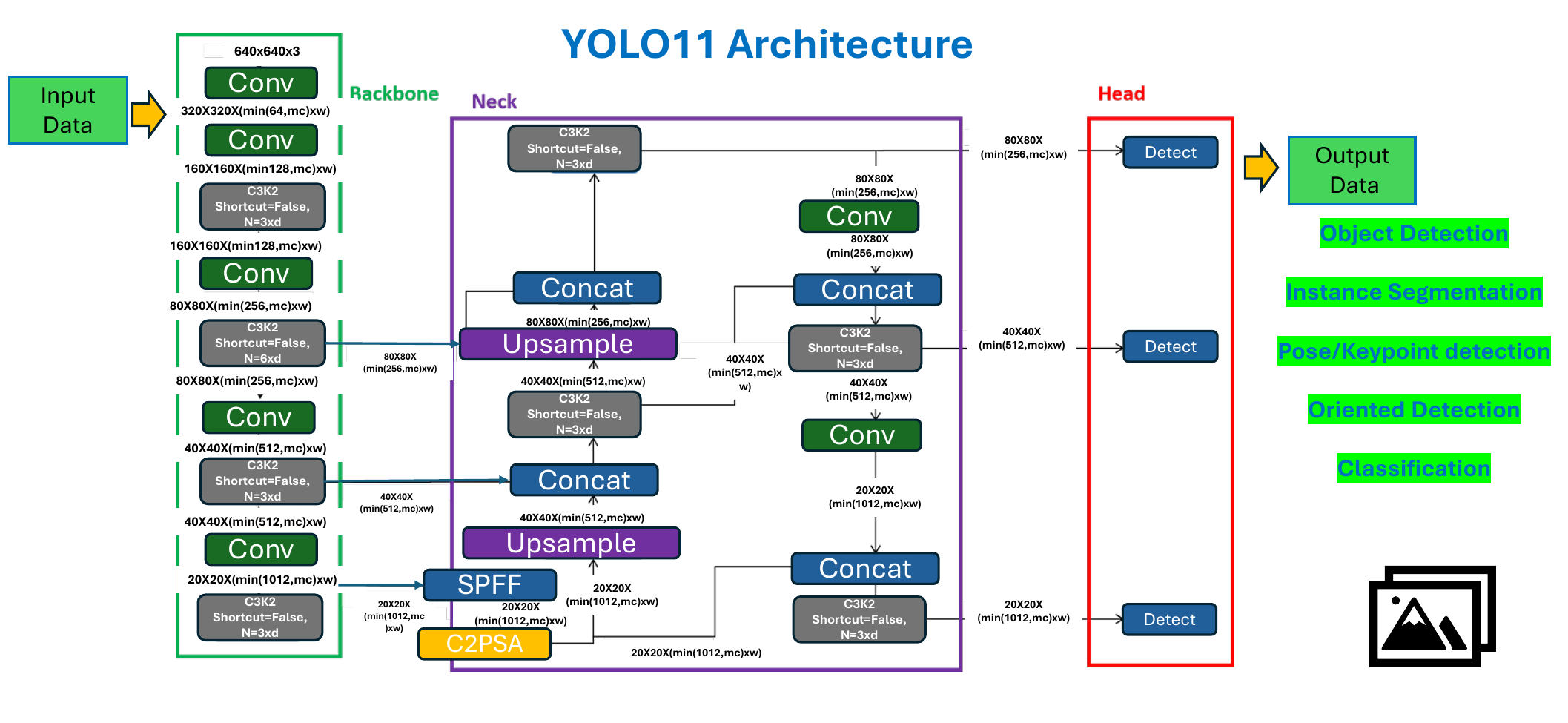}
    \caption{YOLO11 architecture diagram}
    \label{fig:yolov11}
\end{figure}

YOLO11 architecture as illustrated in Figure \ref{fig:yolov11} replaced C2f with compact C3k2 blocks, retained SPPF for multi-scale features, and introduced C2PSA to enhance spatial attention, yielding a leaner backbone/neck with improved feature selectivity. Empirical summaries note fewer parameters than comparable YOLOv8 variants and modest inference speed gains over YOLOv10, with support for a wide task set (detection, segmentation, pose, classification, OBB). The design goal was to preserve throughput while boosting representation in cluttered, small-object scenes across edge and GPU stacks. 

\subsection{YOLOv8: Decoupled Head, Anchor-Free Predictions}
YOLOv8 delivered a more principled detector head and a modernized representation pipeline. The decoupled head separates classification and regression branches, mitigating gradient interference between localization and recognition objectives \cite{fang2024lightweight}. This design reduces optimization entanglement improving convergence smoothness and yielding better calibration at a given operating point (e.g., at fixed mAP or recall). Complementing this head, YOLOv8 formalized an anchor-free assignment strategy: instead of regressing offsets around predefined anchor priors, the model predicts normalized box parameters and confidences directly for candidate points on the feature map \cite{fang2024lightweight, liang2025vst, dolatyabi2025deep}. Anchor-free heads simplify hyperparameter tuning (no anchor clustering) and improve generalization across datasets with differing object aspect-ratio statistics \cite{liu2024using, zhang2024dsp}.

\begin{figure}[h!]
     \centering
     \includegraphics[width=0.99\linewidth]{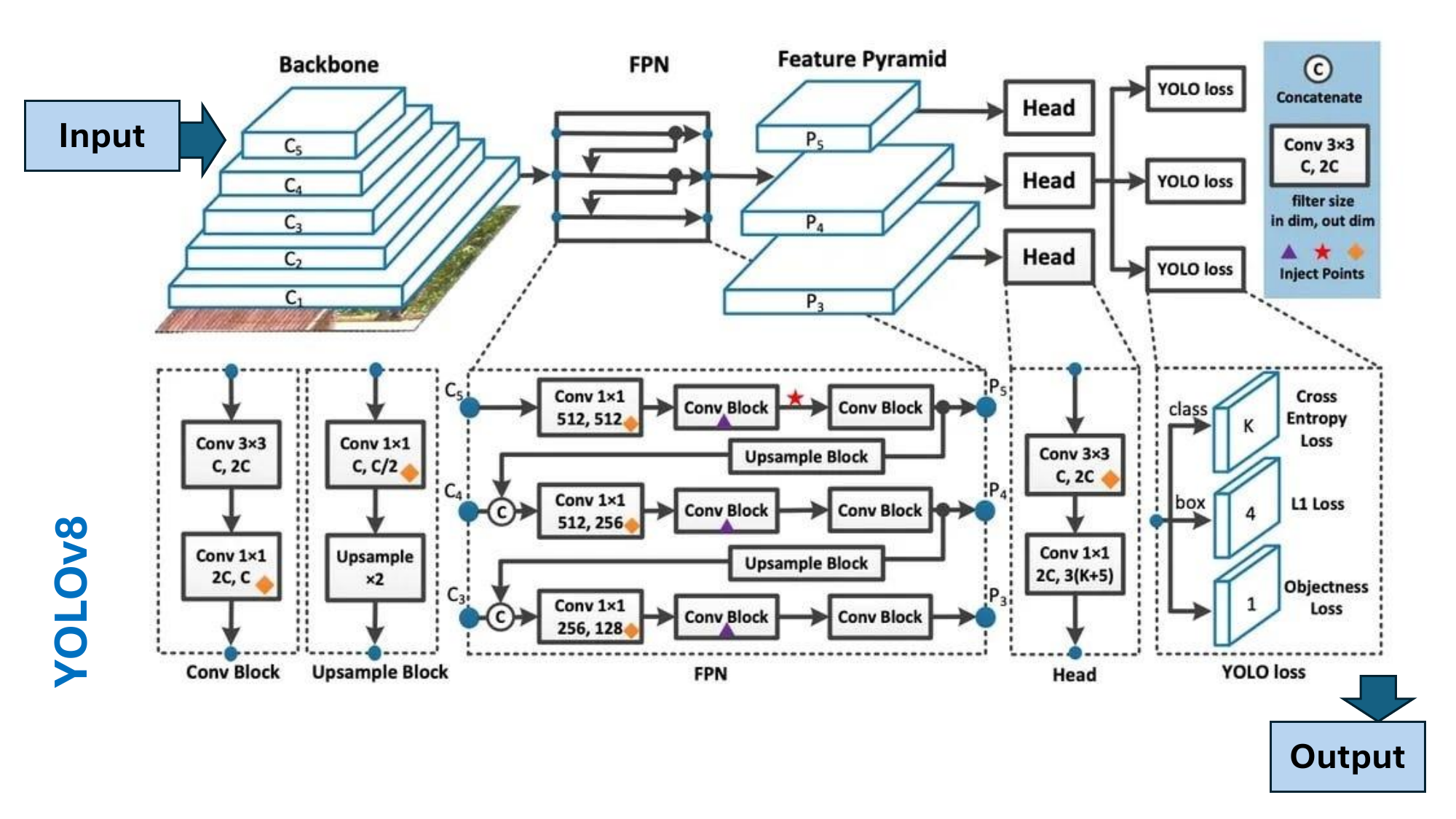}
    \caption{YOLOv8 architecture diagram}
    \label{fig:yolov8}
\end{figure}

The backbone/neck stack evolved toward lighter, high-throughput modules (e.g., C2f-style blocks) that maintain receptive-field richness while reducing memory bandwidth pressure \cite{wang2023bl, yang2023lightweight}. Multi-scale features are fused with attention to stride alignment and aliasing minimization, preserving crisp spatial information critical to small objects. Training refinements included improved label assignment heuristics, class-imbalance mitigation, and stronger regularization under aggressive augmentation, which together stabilized long-horizon training and reduced sensitivity to batch size. These choices enabled coherent scaling across nano to extra-large variants without pathological underfitting or collapse at the smallest scales.

Operationally, YOLOv8 expanded first-class support beyond detection to instance segmentation, classification, and (via an extended head) keypoints/pose. The PyTorch-first design continued to emphasize exportability (ONNX, TensorRT, CoreML, TFLite) and FP16 inference, further smoothing the path from research to deployment. While YOLOv8 still used conventional NMS and (in many settings) distributional regression for boxes, its head decoupling and anchor-free philosophy foreshadowed the end-to-end direction later realized in YOLO26.

The YOLOv8 architecture (Figure \ref{fig:yolov8}) improved convergence and small-object fidelity, complemented by CIoU/DFL losses and an expanded task portfolio (detection, segmentation, pose, tracking, classification). Reported results highlight strong COCO AP with high throughput on TensorRT, and multiple model scales (n–x) for flexible deployment. Collectively, these choices advanced accuracy without sacrificing real-time performance. 

\subsection{YOLOv5: Modularity and PyTorch Adoption}
Ultralytics’ YOLOv5 marked a decisive shift from Darknet to a fully PyTorch-native implementation, catalyzing adoption by unifying training utilities, data pipelines, augmentation strategies, and export tooling under a single, accessible codebase \cite{jocher2022ultralytics}. Architecturally, YOLOv5 popularized a family of depth/width–scaled variants (n/s/m/l/x) built on CSP-inspired backbones and PANet-style necks to improve multi-scale aggregation with minimal latency overhead \cite{ren2022yolov5, dang2023accelerated}. The transition to SiLU activations and careful normalization choices reduced optimization pathologies (e.g., vanishing gradients in deeper stacks) while preserving high-frequency detail important for small-object detection \cite{jocher2021ultralytics}.

A key contribution of YOLOv5 is modularity: components (backbone blocks, neck layers, heads, losses) can be swapped, ablated, or extended with minimal refactoring \cite{johnson2025yolo}. This modularity underpins rapid task adaptation e.g., adding a segmentation head or a classification head without destabilizing the base detector. On the training side, YOLOv5 systematized strong, production-friendly augmentations (mosaic, copy-paste, HSV jitter, random perspective) and reliable schedules (cosine, one-cycle), paired with EMA weight tracking for robust final checkpoints. The data loader and caching logic were engineered for throughput on commodity hardware, enabling high iteration rates that accelerate experimentation cycles.

From a deployment perspective, YOLOv5 leaned into broad format support (ONNX, TensorRT, CoreML), mixed-precision inference (FP16), and lightweight post-processing \cite{nnadozie2024simplifying, fang2021deployment, yan2021real}. Although it retained NMS and conventional bounding-box regression (with optional DFL in later forks), the streamlined graph and conservative operator set made it particularly amenable to embedded and cloud accelerators. Collectively, these design choices positioned YOLOv5 as a practical baseline: not the most novel architecturally, but extremely effective in closing the gap between research-grade code and production-grade inference at scale.

\begin{figure}[h!]
     \centering
     \includegraphics[width=0.89\linewidth]{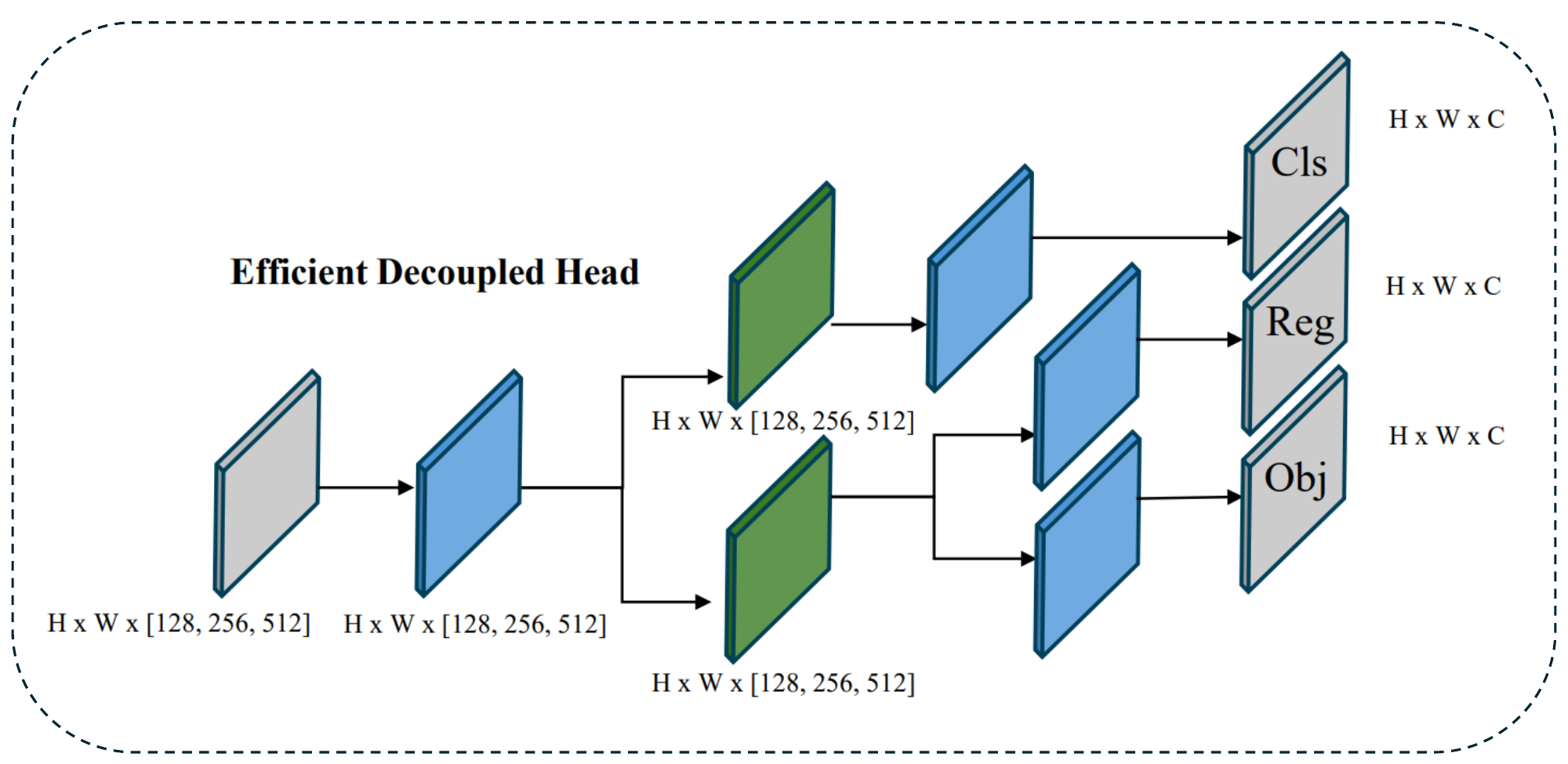}
    \caption{YOLOv5 architecture diagram}
    \label{fig:yolov5}
\end{figure}

The YOLOv5 architecture (Figure \ref{fig:yolov5}) release prioritized reproducibility, streamlined augmentation pipelines, and frictionless export, which accelerated adoption in research and production. The emphasis on developer ergonomics and deployment breadth rather than radical new blocks positioned YOLOv5 as a durable baseline that subsequent versions iterated upon for accuracy/latency trade-offs. 

\section{Benchmarking and Comparative Analysis}

The performance of real-time object detectors is typically measured on the MS~COCO dataset, where the mean Average Precision (mAP) across multiple intersection-over-union thresholds ($0.50$ to $0.95$) quantifies detection quality, and per-image inference latency assesses suitability for edge deployment. This section compares four Ultralytics releases YOLOv5, YOLOv8, YOLO11, and YOLO26 and situates them within the broader landscape by contrasting with YOLOv12, YOLOv13, RT-DETR variants, and the DEIM training framework. Three subsections examine: (i) quantitative evaluation of Ultralytics models on MS~COCO; (ii) cross-comparisons with recent transformer-oriented detectors; and (iii) the metrics used for these evaluations.

\subsection{Quantitative evaluation of YOLOv5, YOLOv8, YOLO11, and YOLO26}

The Ultralytics lineage exhibits steady gains in accuracy and efficiency across successive releases (Refer to Table \ref{tab:ultra_bench}). \textbf{YOLOv5} (2020) introduced a modular PyTorch implementation supporting depth/width-scaled variants (n, s, m, l, x). The SiLU-activated backbone and PANet neck yield competitive accuracy; however, reliance on Non-Maximum Suppression (NMS) and, in some configurations, Distribution Focal Loss (DFL) adds post-processing overhead. On COCO val, YOLOv5u-n (640~px) reports mAP $\approx$34.3\% at $\sim$73.6\,ms per image on CPU (ONNX), while YOLOv5u-s reaches $\approx$43.0\% mAP at $\sim$120.7\,ms. These figures provide a practical baseline for later releases.

\textbf{YOLOv8} (2023) improves convergence and small-object fidelity via a decoupled head and anchor-free design. A C2f backbone with SPPF reduces parameters while preserving receptive-field richness. YOLOv8n (640~px) delivers $\approx$37.3\% mAP at $\sim$80.4\,ms per image on CPU, with larger variants (s, m, l) trading latency for accuracy (e.g., YOLOv8s $\approx$44.9\% mAP at $\sim$128.4\,ms; YOLOv8m $\approx$50.5\% mAP at $\sim$197.5\,ms). Detection, instance segmentation, and classification are first-class tasks, though final predictions remain NMS-based.

\textbf{YOLO11} (2024) targets efficiency and small-object performance with compact C3k2 bottlenecks and a C2PSA attention module for robust feature aggregation. YOLO11n (640~px) attains $\approx$39.5\% mAP with lower CPU latency ($\sim$56.1\,ms) than YOLOv5u-n, while YOLO11s and YOLO11m achieve $\approx$47.0\% and $\approx$50.3\% mAP at $\sim$90.0\,ms and $\sim$171.0\,ms, respectively. The family extends support to pose estimation and oriented bounding boxes, yet still employs NMS at inference.

\textbf{YOLO26} (2025) implements the most radical simplifications: removal of DFL and native end-to-end (NMS-free) inference, alongside Progressive Loss Balancing (ProgLoss), Small-Target-Aware Label Assignment (STAL), and a MuSGD optimizer for stable, fast convergence. YOLO26n (640~px) reports $\approx$39.8\% mAP (up to $\approx$40.3\% in end-to-end mode) at $\sim$38.9\,ms on CPU substantially faster than YOLO11n at comparable accuracy. YOLO26s and YOLO26m reach $\approx$47.2\% and $\approx$51.5\% mAP at $\sim$87.2\,ms and $\sim$220.0\,ms; YOLO26l improves to $\approx$53.0–53.4\% mAP at $\sim$286.2\,ms. These results underscore a deployment-centric design: by eliminating NMS and DFL, CPU inference improves markedly while preserving accuracy across scales.

\begin{table}[h!]
\centering
\caption{Detection performance of Ultralytics YOLO models on MS~COCO validation. Latency is milliseconds per image on CPU (ONNX) and T4 GPU (TensorRT). Variants: n (nano), s (small), m (medium), l (large).}
\label{tab:ultra_bench}
\scriptsize
\begin{tabular}{lcccccc}
\toprule
\textbf{Model} & \textbf{Variant} & \textbf{mAP50–95 (\%)} & \textbf{CPU ONNX (ms)} & \textbf{T4 TensorRT (ms)} & \textbf{Params (M)} & \textbf{FLOPs (B)} \\ \midrule
YOLOv5u & n & 34.3 & 73.6 & 1.06 & 2.6 & 7.7\\
 & s & 43.0 & 120.7 & 1.78 & 7.2 & 17.0\\
YOLOv8 & n & 37.3 & 80.4 & 0.99 & 3.2 & 8.7\\
 & s & 44.9 & 128.4 & 1.90 & 11.2 & 25.9\\
 & m & 50.5 & 197.5 & 3.30 & 25.9 & 63.3\\
YOLO11 & n & 39.5 & 56.1 & 1.5 & 2.6 & 6.5\\
 & s & 47.0 & 90.0 & 1.9 & 9.2 & 16.7\\
 & m & 50.3 & 171.0 & 3.0 & 20.0 & 44.8\\
 & l & 52.2 & 308.7 & 4.8 & 24.8 & 53.4\\
YOLO26 & n & 39.8 (40.3 e2e) & 38.9 & 1.7 & 2.4 & 5.4\\
 & s & 47.2 (47.6 e2e) & 87.2 & 2.7 & 9.5 & 20.7\\
 & m & 51.5 (51.7 e2e) & 220.0 & 4.9 & 20.4 & 68.2\\
 & l & 53.0–53.4 & 286.2 & 6.5 & 24.8 & 86.4\\
\bottomrule
\end{tabular}
\end{table}

\subsection{Cross-comparisons with YOLOv12, YOLOv13, RT-DETR, and DEIM}

Outside the Ultralytics line, several detectors pursue higher accuracy via attention and transformer mechanisms (Refer to Table \ref{tab:other_bench}). \textbf{YOLOv12} introduces efficient area attention and R-ELAN blocks, publishing a spectrum of COCO results from the nano through large variants (e.g., YOLO12n $\sim$40.6\% AP at $\sim$1.64\,ms on T4; YOLO12s $\sim$48.0\% at $\sim$2.61\,ms; YOLO12m $\sim$51.9\% at $\sim$5.73\,ms; YOLO12l $\sim$53.2\% at $\sim$6.14\,ms). \textbf{YOLOv13} advances global context modeling with hypergraph correlation and pipeline-wide aggregation, reporting nano through large tiers that close the gap with transformer detectors (e.g., small and large variants at $\sim$52\% and $\gtrsim$56\% AP with mid-tens of milliseconds on T4). These families emphasize richer feature interactions while retaining convolutional efficiency.

Transformer-style, end-to-end detectors provide a complementary perspective. \textbf{RT-DETR} replaces NMS with bipartite matching, achieving $\sim$51 AP on COCO within a real-time latency envelope on T4; iterative refinements in RT-DETRv2 and RT-DETRv3 push accuracy into the mid-50s AP at modest increases in latency. \textbf{DEIM} (a training framework) further elevates DETR-like systems; combined with D-FINE, published results include AP $\sim$54.7 at $\sim$124\,fps (L) and AP $\sim$56.5 at $\sim$78\,fps (X) on T4. While these transformer pipelines achieve the highest AP, they typically incur higher computational complexity and are less favorable for CPU-bound edge scenarios. In contrast, YOLO26 prioritizes export simplicity, quantization robustness, and NMS-free decoding to reduce latency variance and integration effort on heterogeneous hardware.

\begin{table}[h!]
\centering
\caption{Comparative MS~COCO benchmarking for recent non-Ultralytics YOLO variants and transformer-style detectors. AP is mAP@50–95; latency is per-image on an NVIDIA T4 (TensorRT where applicable).}
\label{tab:other_bench}
\scriptsize
\begin{tabular}{lcccc}
\toprule
\textbf{Model} & \textbf{Variant} & \textbf{AP (\%)} & \textbf{Latency (ms)} & \textbf{Notes} \\ \midrule
YOLOv12 & n / s / m / l & 40.6 / 48.0 / 51.9 / 53.2 & 1.64 / 2.61 / 5.73 / 6.14 & Efficient area attention, R-ELAN \\
YOLOv13 & n / s / l & $\sim$42 / $\sim$52 / $\gtrsim$56 & $\sim$41 / $\sim$48 / $\sim$67 & Hypergraph correlation, pipeline aggregation \\
RT-DETR & base & $\sim$51 & $\sim$46 & End-to-end bipartite matching \\
RT-DETRv2 & base & $\sim$52.7 & $\sim$48 & Enhanced fusion and decoding \\
RT-DETRv3 & base & $\sim$54.7 & $\sim$54 & Further architectural refinements \\
DEIM + RT-DETRv2 & base & $\sim$53.2 &   & Training framework improves AP \\
DEIM-D-FINE & L / X & $\sim$54.7 / $\sim$56.5 & $\sim$8.1 / $\sim$12.8 & $\sim$124 fps / $\sim$78 fps (T4) \\
\bottomrule
\end{tabular}
\end{table}

\subsection{Metrics: precision, recall, F1, mAP, inference speed}

Quantitative evaluation relies on complementary criteria. \textbf{Precision} measures the fraction of predicted boxes that match ground truth, while \textbf{recall} quantifies the fraction of ground-truth objects that are detected. The \textbf{F1 score} the harmonic mean of precision and recall summarizes the trade-off between false positives and false negatives. In object detection, the field predominantly reports \textbf{mean Average Precision} (mAP), computed by integrating precision–recall curves across multiple IoU thresholds and classes. MS~COCO averages AP at IoUs 0.50:0.05:0.95 (AP50–95), a stringent measure that is sensitive to localization accuracy and class balance.

The interplay between accuracy and latency is central to model selection. Smaller models (e.g., nano and small variants) offer favorable throughput on CPUs and embedded GPUs, making them attractive for robotics, UAVs, and smart cameras; larger variants maximize mAP at the expense of inference time, which can be acceptable for server-side analytics. End-to-end designs that remove NMS reduce latency variance and eliminate threshold-tuning overhead in production. In parallel, \textit{exportability} (ONNX, TensorRT, CoreML, TFLite) and \textit{quantization robustness} (FP16/INT8) increasingly determine real-world viability, as deployment often spans heterogeneous accelerators and software stacks. Within this context, the Ultralytics family illustrates a pragmatic trajectory: YOLOv5 establishes a reproducible baseline, YOLOv8 improves learning dynamics and task breadth, YOLO11 strengthens efficiency and small-object fidelity, and YOLO26 consolidates these trends with DFL removal, NMS-free decoding, and quantization-friendly graphs that scale from cloud GPUs to edge devices.

\section{Deployment and Application Perspectives}

The YOLO family is designed not only for research but also for practical deployment across a wide range of hardware platforms and industry applications. This section discusses how YOLOv5, YOLOv8, YOLO11, and YOLO26 are exported, quantized, and deployed on edge devices, and it surveys representative use cases in robotics, agriculture, surveillance, and manufacturing.

\subsection{Export formats and hardware compatibility}

Modern YOLO implementations emphasise portability \cite{munteanu2022sea, ali2024yolo, ray2025llmyoloedge}. The Ultralytics ecosystem provides export scripts that convert PyTorch checkpoints into multiple deployment formats, enabling efficient inference on diverse hardware \cite{jocher2022ultralytics}. YOLOv5 introduced a comprehensive export pipeline: models can be saved as native PyTorch weights (.pt), TorchScript (.torchscript) for embedding in C++ applications, ONNX (.onnx) for cross‑framework interoperability, OpenVINO for Intel CPUs and VPUs, TensorRT (.engine) for NVIDIA GPUs, CoreML (.mlmodel) for iOS devices, TensorFlow SavedModel, GraphDef, and TFLite (.tflite) for Android and microcontroller deployment, as well as specialised formats for Edge TPUs, TensorFlow.js, and PaddlePaddle \cite{jocher2020ultralytics, kim2022object}. Export to ONNX or OpenVINO typically yields up to three‑fold acceleration on CPUs, while conversion to TensorRT can deliver five‑fold GPU speedups. Table~\ref{tab:export_formats} summarises the major export options and the hardware targets they enable for each Ultralytics release.

\begin{table}[h!]
\centering
\caption{Common export formats and compatible deployment hardware. ``Yes'' indicates first‑class support; parentheses denote community or experimental support. \emph{Edge TPUs} refers to Google Coral and similar accelerators.}
\label{tab:export_formats}
\scriptsize
\begin{tabular}{lcccccccc}
\toprule
\textbf{Model} & \textbf{PyTorch} & \textbf{TorchScript} & \textbf{ONNX / OpenVINO} & \textbf{TensorRT} & \textbf{CoreML} & \textbf{TFLite / TF} & \textbf{Edge TPUs} & \textbf{Other} \\ \midrule
YOLOv5 & Yes & Yes & Yes & Yes & Yes & Yes & (Yes) & Paddle, TF.js \\
YOLOv8 & Yes & Yes & Yes & Yes & Yes & Yes & (Yes) & TF.js, Paddle \\
YOLO11 & Yes & Yes & Yes & Yes & Yes & Yes & (Yes) & TF.js, Paddle \\
YOLO26 & Yes & (Yes) & Yes & Yes & Yes & Yes & Yes & TFLite EdgeTPU, ONNX RT \\
\bottomrule
\end{tabular}
\end{table}

YOLOv8 refined the export interface, offering command‑line flags to select formats and optimise for mixed precision (e.g. FP16). YOLO11 retained this flexibility while adding orientated bounding box support in ONNX and TensorRT exports \cite{sapkota2025yolo}. YOLO26’s simplified head and DFL removal further streamline conversion: without distributional regression or NMS, exported graphs contain fewer custom operations and thus map more cleanly to ONNX, TensorRT, CoreML, and TFLite backends. This improved portability ensures models trained in PyTorch can be embedded into C++ pipelines, mobile applications, or web environments with minimal engineering overhead.

\subsection{Quantization and edge inference}
Quantization converts floating‑point models into lower‑precision representations to reduce memory footprint and accelerate inference \cite{wu2021low, xiao2024neural}. The YOLOv5 export pipeline supports half‑precision (FP16) and 8‑bit integer (INT8) quantization through ONNX Runtime, TensorRT, or TFLite. FP16 often halves memory usage and doubles throughput on compatible GPUs; INT8 can provide further gains on CPUs and NPUs, albeit with a small accuracy drop if calibration is not carefully performed. YOLOv8 inherited these capabilities and integrated mixed‑precision training to ease quantization‑aware deployment. YOLO11 continued to support FP16/INT8 exports while improving post‑quantization accuracy by reducing depthwise convolutions and normalising weight distributions.

YOLO26 emphasises quantization robustness: by eliminating DFL and adopting an NMS‑free decoder, its computational graph avoids custom operations that are difficult to quantize. Experiments show that INT8 exports of YOLO26 retain nearly the same mAP as FP32 versions, enabling smooth deployment on resource‑constrained hardware such as the NVIDIA Jetson Orin and Xavier series, Qualcomm Snapdragon AI accelerators, and ARM CPUs. In addition, YOLO26 supports tensor decompositions and channel pruning, allowing further reductions in model size.

Jetson platforms are a common target for edge inference. When compiled with TensorRT in FP16 mode, YOLOv5n yields sub‑2\,ms latency on a T4 GPU; YOLOv8n and YOLO11n exhibit similarly low latencies with improved accuracy. For INT8 inference on Jetson Xavier NX, YOLOv8n runs at around 18–20\,ms per frame while maintaining high accuracy, whereas YOLO26n can achieve comparable accuracy at lower latency owing to the simplified head. On CPUs, deploying ONNX models with OpenVINO or ONNX Runtime results in significant speedups compared with PyTorch; for example, YOLO26n runs around twice as fast as YOLOv8n at similar accuracy on x86 processors. For mobile devices, TFLite and CoreML exports enable real‑time inference; YOLOv8s and YOLO26s can process 30 frames per second on modern smartphones, offering a compelling solution for on‑device vision.

\subsection{Applications in robotics, agriculture, surveillance and manufacturing}

\paragraph{Robotics} Real‑time object detection is fundamental for autonomous navigation, manipulation, and human–robot interaction \cite{rakhimkul2019autonomous, robinson2023robotic, megalingam2023robot}. YOLOv5 and YOLOv8 have been widely used for robotics tasks such as obstacle avoidance, pallet detection, and warehouse inventory scanning due to their balance of speed and accuracy. YOLO11’s improved small‑object detection aids precise localisation of tools or components. YOLO26’s NMS‑free predictions reduce system latency and simplify integration with downstream planners; its support for pose estimation and oriented bounding boxes enables grasp planning and six‑degree‑of‑freedom pose estimation. On mobile robots with Jetson Orin, a quantized YOLO26n can perform multi‑task perception (detection, segmentation, keypoints) within a 10–20\,ms budget, permitting closed‑loop control at 50–100\,Hz.

\paragraph{Agriculture} In precision agriculture, UAVs and ground robots monitor crop health, detect weeds, and estimate yield. YOLOv5 and YOLOv8 have been applied to count fruit, identify diseases, and segment weeds \cite{badgujar2024agricultural, zoubek2025advancing}. YOLOv8’s instance segmentation head helps separate overlapping leaves, while the classification head differentiates crop species. YOLO11’s oriented detection can handle elongated objects (e.g. vines) and its improved recall benefits small buds or pests \cite{dai2025yolov11}. YOLO26’s STAL mechanism enhances detection of small fruits under occlusion, and its streamlined exports facilitate deployment on low‑power field robots. Quantized models on embedded GPUs allow processing high‑resolution aerial imagery in real time, enabling timely interventions such as targeted spraying or harvesting.

\paragraph{Surveillance} Video surveillance systems require robust detection under varied lighting and crowded scenes \cite{sapkota2025yolo, nguyen2021yolo, mostafa2024yolo}. YOLOv5 and YOLOv8 form the backbone of many security camera products, offering real‑time inference on edge devices with optional cloud off‑loading. YOLO11 increases detection fidelity for small objects, such as distant pedestrians, and its pose estimation module supports behavioural analysis. YOLO26’s NMS‑free output simplifies pipeline design by removing threshold tuning, and its quantization robustness enables deployment on CPUs in network video recorders. Integration with segmentation tasks helps delineate intrusion zones and improve tracking; oriented bounding boxes facilitate vehicle pose estimation in traffic monitoring.

\paragraph{Manufacturing} Automated inspection and assembly lines demand fast and accurate detection of defects and components \cite{pham2023yolo, mao2024automated}. YOLOv5 and YOLOv8 have been used for defect detection in printed circuit boards, quality assessment of food products, and tool presence verification. YOLO11’s efficient architecture and improved small‑object recall suit high‑speed conveyors where millimetre‑scale defects must be captured. Oriented bounding boxes allow proper alignment of irregular parts for robotic pick‑and‑place. YOLO26’s end‑to‑end inference and removal of DFL reduce system latency and memory, enabling deployment on programmable logic controllers and edge accelerators. Its multi‑task capabilities (segmentation and classification) streamline integration with vision‑guided robotic arms for assembly and packaging. Moreover, the simplified graph eases regulatory validation for safety‑critical environments where deterministic execution and explainability are important.

\section{Challenges and Future Directions}
\label{sec:challenges_future}

The evolution from YOLOv5 and YOLOv8 through YOLO11, YOLO26, and YOLO27
demonstrates a continuing transition from conventional dense convolutional
detectors toward increasingly unified, end-to-end, scale-adaptive, and
deployment-aware perception systems. YOLO27 extends this trajectory through
its dual architectural strategy, combining streamlined CNN-based detection
for compact n/s variants with query-based transformer decoding for m/l
variants and deep-stage UltraViT attention in YOLO27l. Nevertheless, several
fundamental challenges remain unresolved. Dense and occluded scenes, domain
shift, architectural adaptation across computational budgets, open-vocabulary
recognition, multimodal vision--language understanding, temporal reasoning,
and hardware-aware optimization represent particularly important research
directions. Addressing these limitations could transform future YOLO systems
from task-specific real-time visual detectors into more general,
language-conditioned and context-aware perception engines for robotics,
autonomous systems, edge AI, and interactive computer vision.

\subsection{Dense Scenes, Occlusion, and Domain Generalization}

Dense environments remain challenging for real-time object detectors because
multiple instances may overlap spatially, share similar visual
characteristics, or occupy only a small number of pixels. Conventional
Non-Maximum Suppression (NMS) can incorrectly eliminate valid detections when
objects exhibit substantial overlap, producing missed detections or false
positives \cite{hosain2024synchronizing,nikouei2025small,
shao2025context}. Methods such as Soft-NMS \cite{bodla2017soft}, adaptive
box-fusion strategies \cite{mees2016choosing}, and weighted suppression
\cite{wang2021domain,zhao2020suppress} alleviate some of these problems, but
often introduce additional hyperparameters and post-processing complexity.
Anchor-free detectors can improve localization flexibility
\cite{xin2021pafnet,tang2025gpdet}, although severe occlusion, crowded
small-object distributions, and visually similar neighboring instances remain
difficult.

YOLO26 and YOLO27 partially alter this problem by reducing dependence on
traditional NMS. In particular, YOLO27m/l employ query-based decoding and
native NMS-free inference, potentially providing a more natural mechanism for
reasoning over competing object hypotheses. However, NMS-free prediction does
not itself resolve severe occlusion or ambiguity. Query competition,
duplicate suppression during training, limited query capacity, and assignment
under crowded conditions remain important areas requiring systematic
evaluation. For YOLO27n/s, which can operate through either dense NMS-based or
one-to-one NMS-free prediction, controlled evaluation in dense scenes could
also reveal when each inference strategy is preferable. Future benchmarks
should therefore stratify performance by object density, occlusion ratio,
object size, and inter-object overlap rather than reporting only aggregate
mAP.

Domain generalization represents an equally important limitation. Models
trained on curated datasets can experience substantial performance degradation
when deployed under unseen illumination, weather, background, geographic,
sensor, or object-appearance conditions
\cite{oza2023unsupervised,singhal2023domain}. Unsupervised domain adaptation,
adversarial feature alignment, cycle-consistent style transfer, and
pseudo-label refinement attempt to reduce this discrepancy, while
meta-learning and continual-learning approaches can support adaptation to new
environments \cite{oza2023unsupervised}. Nevertheless, catastrophic
forgetting, unstable pseudo-labels, and insufficient standardized
cross-domain benchmarks remain unresolved. Future Ultralytics YOLO models
could combine self-supervised pretraining, domain-aware normalization,
uncertainty estimation, continual adaptation, and test-time adaptation to
improve robustness without requiring complete retraining for every deployment
environment.

\subsection{Beyond Fixed Scaling: Adaptive CNN--Transformer and Query Architectures}

The traditional YOLO scaling philosophy largely assumes that nano, small,
medium, large, and extra-large variants share a common architectural template
whose depth, width, and computational capacity are scaled. YOLO27 introduces
an important departure from this assumption. Compact YOLO27n/s models retain
streamlined CNN-based detection, whereas YOLO27m/l employ query-based
transformer decoding, and YOLO27l additionally incorporates deep-stage
self-attention through UltraViT. Consequently, hybrid CNN--Transformer design
should no longer be regarded exclusively as a hypothetical future direction
for YOLO; YOLO27 already provides an initial implementation of
\textit{scale-dependent architectural specialization}.

This design nevertheless opens several unanswered questions. Convolutional
networks efficiently capture local spatial patterns, whereas transformers
provide mechanisms for long-range feature interaction and global contextual
reasoning. Detection Transformers (DETR) demonstrated the potential of
query-based set prediction, although early implementations suffered from slow
optimization and substantial computational requirements. Subsequent
architectures have increasingly combined local convolution with sparse,
hierarchical, or deformable attention. Similarly, vision architectures such
as CoAtNet and Next-ViT demonstrate how convolution and attention can be
distributed across network stages, while mechanisms such as HyperACE explore
higher-order feature relationships.

A logical post-YOLO27 research direction is therefore \textbf{dynamic
architecture routing}. Rather than permanently assigning CNN inference to
small models and query decoding to larger variants, future detectors could
select computational pathways according to scene complexity. Sparse scenes
could activate inexpensive convolutional pathways, whereas crowded,
ambiguous, or context-dependent scenes could dynamically invoke additional
attention layers or query-decoding stages. Adaptive query counts, token
pruning, conditional attention, early exiting, and dynamic input resolution
could further allocate computation according to perceptual difficulty.
Neural architecture search could jointly optimize these decisions against
accuracy, latency, memory, and energy constraints. Such systems would move
YOLO beyond static model scaling toward \textit{input-adaptive computation},
where computational effort is determined not only by the selected model size
but also by the complexity of the observed scene.

\subsection{Open-Vocabulary Detection, Vision--Language Models, and Multimodal YOLO}

Perhaps the most consequential limitation of current YOLO systems is their
predominantly closed-vocabulary perception paradigm. Traditional detectors
learn predefined object categories and consequently have limited ability to
recognize semantic concepts that were absent from their annotated training
sets \cite{perera2019deep,kaul2022label,bodesheim2015local}. Open-vocabulary
detection addresses this limitation by connecting visual representations with
language embeddings learned by large-scale vision--language or foundation
models \cite{wu2024towards}. Models such as CLIP, ALIGN, and BLIP demonstrate
that visual and textual concepts can be represented within semantically
aligned embedding spaces. Open-vocabulary detectors such as DETIC
\cite{zhou2022detecting} and GLIP \cite{li2022grounded} further demonstrate
that language-conditioned representations can support detection beyond a
fixed set of predefined classes.

Although YOLO27 introduces query-based reasoning for its higher-capacity
detectors, its object queries remain fundamentally different from
\textbf{natural-language-conditioned queries}. This distinction identifies a
major opportunity for future YOLO architectures. A multimodal YOLO could
replace or augment the fixed classification vocabulary with a text encoder,
allowing users to specify target concepts using natural-language prompts.
Instead of requesting only predefined categories such as ``person,'' ``car,''
or ``apple,'' a future detector could potentially respond to compositional
instructions such as ``detect the partially occluded green fruit,'' ``locate
the damaged component beside the motor,'' or ``find the red package on the
upper shelf.'' Such capability would transform YOLO from closed-set object
detection toward \textbf{language-grounded real-time perception}.

A potential architecture could combine the efficient YOLO visual backbone
with a lightweight text encoder and cross-modal feature alignment. Image
features and textual tokens could be projected into a shared embedding space,
while language-conditioned object queries interact with multi-scale visual
features through cross-attention. Training could jointly incorporate
bounding-box annotations, referring expressions, image--caption pairs, and
synthetically generated textual descriptions
\cite{zhang2025unified,laina2019towards}. This formulation could support
open-vocabulary detection, referring-object localization, zero-shot transfer,
and prompt-conditioned perception within a common real-time architecture.

A further step would be integration with \textbf{multimodal large language
models (MLLMs)}. YOLO could serve as a high-frequency visual grounding module
while an MLLM provides semantic interpretation and reasoning. The detector
could identify candidate objects and their spatial relationships, while the
multimodal reasoning layer interprets user instructions, resolves ambiguous
references, or determines which objects are relevant to a downstream task.
For robotics, this could create a direct pathway from natural-language
instruction to grounded perception and eventually action. However, such
integration introduces substantial challenges, including hallucination,
language ambiguity, calibration of open-set confidence, synchronization
between detector and language model, increased memory consumption, and the
difficulty of retaining YOLO-level real-time throughput.

\subsection{From Frame-Level Detection to Temporal and Context-Aware Perception}

Most YOLO benchmarks continue to evaluate individual images, whereas many
real-world applications operate on continuous video. Autonomous vehicles,
UAVs, mobile robots, surveillance systems, agricultural machines, and
industrial inspection platforms observe temporally correlated scenes in which
information from previous frames can substantially improve current
interpretation. Conventional frame-by-frame detection repeatedly computes
similar features and can produce unstable predictions when objects become
temporarily occluded, blurred, or partially outside the field of view.

Future YOLO systems could therefore incorporate explicit
\textbf{temporal memory}. Rather than treating tracking as a downstream
association process alone, temporal features could be retained across frames
and integrated directly into detection. Lightweight recurrent memory,
temporal attention, object-query persistence, or state-space representations
could propagate object identity and contextual information through time.
Query-based YOLO27m/l models provide a particularly interesting foundation
for this direction because object queries could potentially persist across
frames, allowing previously detected objects to maintain temporal
representations even during short-term occlusion.

This development would shift YOLO from image-level detection toward
\textit{stateful visual perception}. A temporally persistent model could
reason not only about what is visible in the current frame but also about
where an object was previously observed, whether it is moving, and whether a
temporary disappearance is likely caused by occlusion. Combining temporal
memory with language conditioning could ultimately support queries such as
``track the worker who entered the restricted area'' or ``locate the fruit
that was previously occluded by this branch.'' Achieving these capabilities
while maintaining low latency and bounded memory represents an important
research challenge.

\subsection{Uncertainty, Reliability, and Failure-Aware Detection}

Increasing architectural complexity also creates a need for more explicit
assessment of prediction reliability. Conventional YOLO outputs provide
classification confidence and localization estimates, but these values do not
necessarily constitute calibrated measures of epistemic or aleatoric
uncertainty. This limitation becomes particularly important under domain
shift, severe weather, sensor degradation, adversarial conditions, or
open-vocabulary deployment, where high-confidence predictions may still be
incorrect.

Future YOLO architectures could therefore integrate uncertainty estimation,
out-of-distribution detection, and selective prediction. Rather than always
returning a detection, the model could identify situations in which visual
evidence is insufficient and trigger an alternative computational pathway.
For example, uncertain CNN predictions could activate a higher-capacity
query-based decoder, request temporal evidence from neighboring frames, or
invoke a multimodal reasoning model. Such \textbf{failure-aware inference}
would be particularly valuable in safety-critical robotics and autonomous
systems because computational resources could be concentrated on uncertain
cases rather than uniformly increasing model complexity.

Evaluation protocols should similarly move beyond aggregate precision,
recall, and mAP. Calibration error, uncertainty under domain shift,
out-of-distribution rejection, robustness to corruptions, failure-recovery
rate, and confidence-conditioned accuracy would provide a more complete
assessment of real-world reliability. This becomes increasingly important if
future YOLO generations transition from passive detection toward
language-conditioned and decision-support perception.

\subsection{Edge-Aware Training and Hardware-in-the-Loop Optimization}

Real-time deployment remains a defining requirement of the YOLO paradigm.
Quantization-aware training, mixed-precision inference, channel pruning, and
neural architecture search can substantially reduce computational and memory
requirements, but optimization based only on FLOPs or parameter count may
poorly predict actual deployment performance. Memory bandwidth, operator
support, kernel-launch overhead, tensor-core utilization, thermal constraints,
and compiler behavior can cause models with similar theoretical complexity to
exhibit substantially different real-world latency.

Hardware-in-the-loop optimization offers a pathway toward direct
\textbf{algorithm--hardware co-design}. Instead of treating latency, power
consumption, and temperature as post-training measurements, these quantities
can become objectives during architecture selection. Once-For-All-style
supernetworks illustrate how a common architecture can be specialized for
different deployment constraints, while reinforcement-learning-based
controllers can dynamically adjust network width, depth, resolution, or
precision. Device-specific operators could similarly be incorporated during
training so that architecture optimization reflects the actual execution
characteristics of GPUs, NPUs, DSPs, and embedded accelerators.

YOLO27's scale-dependent architecture makes this direction particularly
relevant. The decision to retain efficient CNN computation for n/s models
while introducing query decoding for m/l models already reflects an implicit
relationship between architecture and computational budget. Future
Ultralytics systems could make this relationship explicit by automatically
selecting CNN, attention, query, precision, and resolution configurations
according to the target device. A battery-powered UAV, for example, could
dynamically reduce precision or resolution under energy constraints, whereas
an autonomous robot connected to a high-performance GPU could activate
additional queries or attention layers for difficult scenes. Continual
profiling and adaptive quantization could further maintain efficiency as
hardware conditions or workload characteristics change.

\subsection{A Post-YOLO27 Roadmap: Toward Multimodal, Adaptive, and Embodied Perception}

Taken together, the progression from YOLOv5 to YOLO27 suggests that future
development may be driven less by simply increasing network depth or
parameter count and more by expanding the \textbf{semantic, temporal, and
adaptive capabilities} of real-time perception. YOLOv8 established
anchor-free prediction within the modern Ultralytics ecosystem, YOLO11
strengthened efficiency and multi-task representation, YOLO26 advanced
end-to-end NMS-free deployment, and YOLO27 introduces scale-adaptive
specialization between CNN and query-based architectures. The next major
research frontier is therefore likely to involve capabilities that remain
largely outside conventional closed-vocabulary, frame-level object detection.

A potential post-YOLO27 roadmap can be conceptualized across four interacting
directions. First, \textbf{multimodal perception} could introduce
vision--language alignment, natural-language prompts, referring expressions,
and open-vocabulary detection. Second, \textbf{temporal intelligence} could
introduce persistent object queries and visual memory for video and robotic
perception. Third, \textbf{adaptive computation} could dynamically route
inputs through CNN, attention, or query-based pathways according to scene
complexity, uncertainty, and available hardware. Fourth,
\textbf{embodied integration} could connect detection outputs to downstream
planning and action, allowing robots and autonomous systems to use grounded
visual concepts as part of closed-loop decision-making.

Importantly, these directions should not be interpreted as announced features
of future Ultralytics releases, but as research opportunities motivated by
gaps in the current YOLO paradigm. Their practical realization will require
maintaining the property that originally distinguished YOLO: real-time
performance. Adding large language encoders, temporal memory, multimodal
cross-attention, or extensive transformer reasoning without controlling
computational cost could undermine the latency advantages of the architecture.
The central challenge for future YOLO research is therefore not simply to
increase model intelligence, but to determine how \textbf{open-vocabulary
semantics, multimodal reasoning, temporal memory, uncertainty awareness, and
adaptive computation can be incorporated while preserving real-time
operation}. Solving this problem could move YOLO beyond ``You Only Look
Once'' as a frame-level detector toward a broader real-time perception
framework capable of seeing, grounding, remembering, and contextualizing
visual information across edge, robotic, and autonomous environments.

\section*{Acknowledgment}
This work was supported in part by the National Science Foundation (NSF) and the United States Department of Agriculture (USDA), National Institute of Food and Agriculture (NIFA), through the “Artificial Intelligence (AI) Institute for Agriculture” program under Award Numbers AWD003473 and AWD004595, and USDA-NIFA Accession Number 1029004 for the project titled “Robotic Blossom Thinning with Soft Manipulators.” Additional support was provided through USDA-NIFA Grant Number 2024-67022-41788, Accession Number 1031712, under the project “ExPanding UCF AI Research To Novel Agricultural Engineering Applications (PARTNER).”

\bibliographystyle{unsrt}  
\bibliography{references}

\end{document}